# AirPilot: Interpretable PPO-based DRL Auto-Tuned Nonlinear PID Drone Controller for Robust Autonomous Flights

Junyang Zhang
Department of EECS & CS
University of California, Irvine
junyanz9@uci.edu

Cristian Emanuel Ocampo Rivera
Department of CS
University of California, Irvine
cocampor@uci.edu

Kyle Tyni
Department of EECS
University of California, Irvine
ktyni@uci.edu

Steven Nguyen
Department of EECS
University of California, Irvine
stevn10@uci.edu

**Abstract** – Navigation precision, speed and stability are crucial for safe Unmanned Aerial Vehicle (UAV) flight maneuvers and effective flight mission executions in dynamic environments. Different flight missions may have varying objectives, such as minimizing energy consumption, achieving precise positioning, or maximizing speed. A controller that can adapt to different objectives on the fly is highly valuable. Proportional Integral Derivative (PID) controllers are one of the most popular and widely used control algorithms for drones and other control systems, but their linear control algorithm fails to capture the nonlinear nature of the dynamic wind conditions and complex drone system. Manually tuning the PID gains for various missions can be time-consuming and requires significant expertise. This paper aims to revolutionize drone flight control by presenting the AirPilot, a nonlinear Deep Reinforcement Learning (DRL) - enhanced Proportional Integral Derivative (PID) drone controller using Proximal Policy Optimization (PPO). AirPilot controller combines the simplicity and effectiveness of traditional PID control with the adaptability, learning capability, and optimization potential of DRL. This makes it better suited for modern drone applications where the environment is dynamic, and mission-specific performance demands are high. We employed a COEX Clover autonomous drone for training the DRL agent within the simulator and implemented it in a real-world lab setting, which marks a significant milestone as one of the first attempts to apply a DRL-based flight controller on an actual drone. Airpilot is capable of reducing the navigation error of the default PX4 PID position controller by 90%, improving effective navigation speed of a fine-tuned PID controller by 21%, reducing settling time and overshoot by 17% and 16% respectively. Notably, adding a $20,000 indoor Vicon tracking system offers <1mm positioning accuracy. To navigate the drone in the shortest collision-free trajectory, we also built a three-dimensional A* path planner and implemented it into the real flight.

## I. INTRODUCTION:

The proliferation of Unmanned Aerial Vehicles (UAVs) in various applications, ranging from surveillance to delivery services, has underscored different needs for precise, fast, and stable navigation. As UAVs increasingly operate in dynamic and unpredictable environments, the demands on reliable task-specific control systems have grown significantly. Traditional Proportional Integral Derivative (PID) controllers have long been a staple in UAV control due to their simplicity and effectiveness in a wide range of conditions [1]. However, the linear nature of PID controllers often falls short in addressing the complex, nonlinear dynamics inherent to UAVs, particularly when navigating through turbulent wind conditions or executing intricate maneuvers. Additionally, tuning PID controllers for optimal performance can be a challenging and time-consuming process. While heuristic methods like Ziegler-Nichols [2] and Cohen-Coon [3] have provided systematic ways to tune PID controllers, they often require subsequent fine-tuning and may not be ideal for all systems and flight tasks, especially those that are nonlinear or have time-varying dynamics.

This limitation necessitates the exploration of more advanced control strategies that can adapt to the evolving conditions of the UAV's operational environments. Recent advancements in machine learning, particularly in Deep Reinforcement Learning (DRL), present a promising solution. DRL algorithms have shown great potential in optimizing control systems by enabling them to learn and adapt from their interactions with the environment. By enhancing traditional PID controllers with DRL, specifically through the integration of Proximal Policy Optimization (PPO) [4], we can create an adaptive control system that not only maintains the simplicity of PID control but also leverages the learning and optimization capabilities of DRL.

In this paper, we propose a novel approach to UAV control by implementing a DRL-enhanced adaptive PID controller, which we call AirPilot. This controller is designed to improve the navigation precision, speed, and stability of UAVs in complex and dynamic environments. Our method is validated using a COEX Clover 4 autonomous drone, which is first trained in a simulated environment using the Gazebo simulator and subsequently deployed in the UCI Resilient Cyber Physical System Laboratory. More importantly, our novel DRL controller directly bypasses the PX4 autopilot position PID controller and does not require any modification to the original flight controller, so it can be seamlessly integrated with any drone utilizing the PX4 autopilot flight control software, as we will demonstrate in the subsequent sections. To further enhance the precision of autonomous flight, we integrate a $20,000 indoor Vicon tracking system, which provides sub-millimeter accuracy in positioning, ensuring high precision in flight operations [5].

Moreover, to optimize the UAV's navigation efficiency, we incorporate a three-dimensional A* path planner [6]. This planner is crucial in enabling the drone to navigate the shortest collision-free trajectory, thereby enhancing the overall mission success rate in complex environments.

This study aims to demonstrate the superiority of the **AirPilot** controller over traditional linear PID controllers in modern UAV applications, providing a robust solution for the increasingly demanding requirements of autonomous flight in dynamic and unpredictable environments.

## II. RELATED WORK

### A. Proximal Policy Optimization

Fig.1 below shows a typical proximal policy optimization algorithm. Each iteration, each of N actors (N=1 in our case) collect T timesteps of data as a batch. Then we calculate the surrogate loss $L$ on these NT timesteps of data and maximize it with Adam optimizer for K epochs. As an on-policy algorithm, PPO learns directly from the data generated by the current policy. Unlike off-policy algorithms, which store experiences in a replay buffer and can use data from previous policies, on-policy algorithms typically discard data after a single update. However, PPO allows for multiple updates (or epochs) using the same batch of collected data, making it more sample-efficient compared to other DRL algorithms. Thus, it requires fewer interactions with the environment to learn an effective policy. This efficiency is crucial for drone controllers, where rapid learning and adaptation are necessary to maintain high performance.

```
for iteration=1,2,... do
    for actor=1,2,...,N do
        Run policy π_{θ_old} in environment for T timesteps
        Compute advantage estimates Â_1,...,Â_T
    end for
    Optimize surrogate L wrt θ, with K epochs and minibatch size M ≤ NT
    θ_old ← θ
end for
```

Figure 1. Pseudo code of the PPO algorithm.

As shown in [4], the objective function of the Proximal Policy Optimization includes the clipped policy objective (See Eq.1), the value function loss (See Eq.2), and the entropy bonus (See Eq.3). The clipped objective function $L^{CLIP}(\theta)$ measures the improvement of the policy $\theta$ compared to the old policy $\theta_{old}$ and it is designed for PPO to make stable and gradual policy updates. This prevents large, destabilizing changes to the controller's policy, ensuring that the PPO-based PID controller can adapt to new conditions without causing erratic behavior or instability in the drone's flight. The objective function $L_t^{CLIP+VF+S}(\theta)$ is defined as follow for the optimizer to maximize:

$$L^{CLIP}(\theta) = E_t[min(r_t(\theta)A_t, clip(r_t(\theta), 1-\epsilon, 1+\epsilon)A_t)] \quad (1)$$
$$L_t^{VF}(\theta) = (V_\theta(s_t) - V_t^{targ})^2 \quad (2)$$
$$L_t^{CLIP+VF+S}(\theta) = E_t[L_t^{CLIP}(\theta) - c_1 L_t^{VF}(\theta) + c_2 S[\pi_\theta](s_t)] \quad (3)$$

Where $\epsilon$ is a hyperparameter, say $\epsilon=0.2$. $r_t(\theta)$ is a ratio that specifies how much the new policy $\theta$ is changed with respect to the old policy $\theta_{old}$. $A_t$ is the advantage of each state-action pair, and it denotes how much better or worse a particular action was compared to the average action expected at that state under the old policy. $L_t^{VF}(\theta)$ is a squared-error loss that calculates the difference between the predicted values of the state from the "critic" neural network and its actual values (See Fig.2). To learn the optimal policy, the "actor" network maps observations (states) from the environment to actions. As shown in Fig.2, the "actor" network shares the same parameters with the "critic" network but with a different head, reducing training complexity and making both networks well-aligned. $c_1$ and $c_2$ are coefficients, and $S$ denotes an entropy loss that encourages exploration by maximizing the entropy of the policy's action distribution during training.

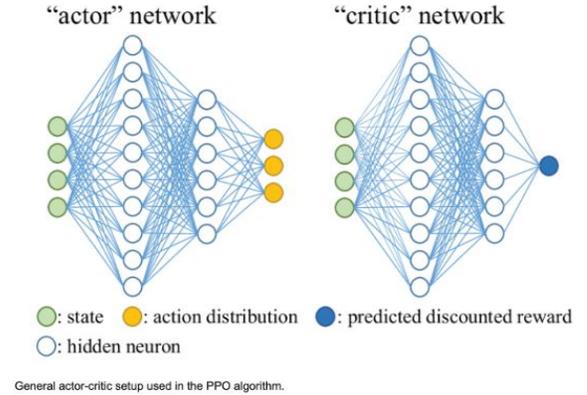

Figure 2. PPO's policy and value networks.

During training, a drone controller needs to balance exploring new actions to improve performance and exploiting known actions that yield good results. Too much exploration can lead to instability, while too much exploitation can cause the controller to miss better solutions. PPO inherently balances exploration and exploitation through its clipped objective and entropy regularization. This ensures that the drone controller remains exploratory enough to adapt to new situations while still exploiting reliable strategies for maintaining stable flight.

### B. Proportional-Integral-Derivative Position Controller

As shown in Eq.5, the PID control law for position controllers is given below. It is a fundamental control strategy in PX4 autopilot flight controllers where the velocity command is determined based on the position error, its derivatives and integrals. The Position Error (PE) is defined in Eq.4 as the absolute distance between the target and the current location.

$$PE = norm(TargetPosition - CurrentPosition) \quad (4)$$

$$Velocity = K_p PE + K_D \frac{dPE}{dt} + K_I \int_t PE \quad (5)$$

Here $K_p$ is the proportional gain that scales the immediate position error, $K_D$ is the derivative gain that scales the rate of change of the position error ($\frac{dPE}{dt}$), providing a damping effect and anticipating future errors, and $K_I$ is the integral gain that scales the accumulated position error ($\int_t PE$) over time, correcting any long-term bias and eliminating steady-state errors. By combining these three terms, the PID controller adjusts the velocity to reduce the position error as efficiently as possible, ensuring accurate and stable positioning of the system. The position controller generates the velocity commands using the given PID control law. Similarly, those commands are fed to the velocity PID controller to generate acceleration setpoints, which is eventually passed to the default PX4-autopilot attitude/rate controllers and mixers.

## III. SETUP:

We carefully selected a range of algorithms, software and hardware to develop a reliable and efficient drone navigation system using Deep Reinforcement Learning (DRL).

### A. Software

The backbone of our coding work was Python 3.1, which we used for scripting the DRL models and managing hardware interactions. To test various software setups efficiently, VMware 17 provided us with the capability to create virtual environments, and Ubuntu 20.04 served as a stable and compatible operating system. We also incorporated the Robot Operating System (ROS) to control drone automation effectively, providing a versatile framework for our software. For simulating the COEX Clover drone's environment and testing its responses, Gazebo Simulator was an indispensable tool, allowing us to trial our models under a variety of conditions (Fig.3). Extended Kalman Filter 2 (EKF2) is used within the standard PX4 Autopilot flight control software for state estimation. The Vicon Tracker Software played a critical role in precisely tracking the drone's movement, providing valuable data for refining our DRL approach (Fig.4).

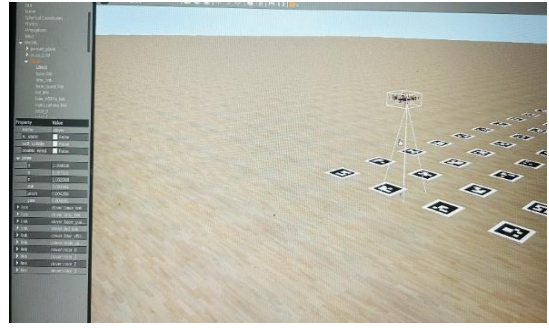

Figure 3. Clover drone in the Gazebo simulator.

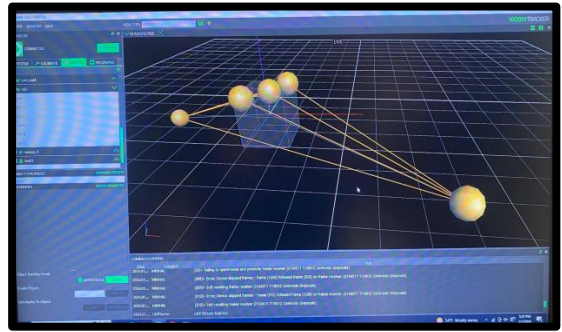

Figure. 4 Vicon Tracker software.

### B. Hardware

On the hardware side, our $355 \times 355 \times 125mm$ COEX Clover 4 autonomous drone was built from essential components like a battery, frame, motors, and propellers, along with necessary assembly materials such as screws and bolts (Fig.5). The Phawk 4 Controller managed the flight control and navigation, while the Raspberry Pi 4 handled onboard computational tasks. A key component in our setup was the 14-Camera Vicon System with IR trackers, used to capture exact positional and orientational data of the drone during flights within 1mm accuracy (Fig.6).

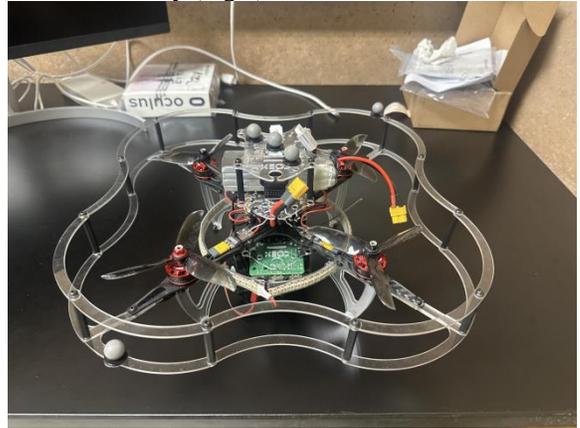

Figure 5. Assembly of the COEX Clover 4 autonomous drone.

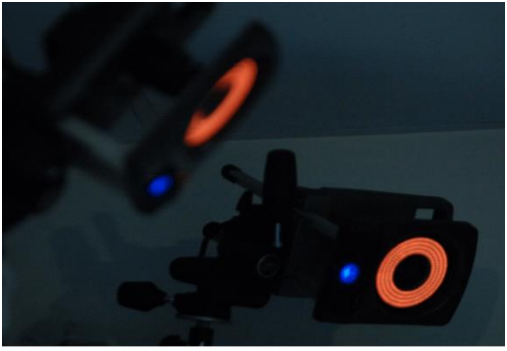

Figure 6. Well-calibrated Vicon Indoor Tracking System consisting of both 14 Vicon cameras.

For software testing and DRL model development, our computers needed to be robust and equipped with a minimum of a 4-core CPU and 8GB of RAM to smoothly run simulations, process data, and train the DRL models. This blend of advanced software and sturdy hardware was crucial in successfully developing, testing, and implementing our DRL-based drone navigation system.

*C. Standards and Protocols*

We also focused on essential standards and protocols to ensure effective and reliable operations. We integrated Wi-Fi (IEEE 802.11) for stable wireless communication, which is crucial for the drone's data exchange with the Vicon system and control interfaces. For direct hardware interactions, particularly between the Raspberry Pi and the flight controller, the USB standard was key for quick and accurate data transfer.

We also implemented specific protocols to enhance performance: USB for rapid internal communication, UDP (User Datagram Protocol) for its fast data transmission between the Vicon system and the drone Raspberry Pi, and MavLink (Micro Air Vehicle Link) to ensure seamless communication between the drone's onboard system and external ground control station which sends out the navigation command [7].

These standards and protocols were selected to optimize the drone's functionality, emphasizing speed, accuracy, and consistency in line with industry norms.

## IV. METHODS:

*A. AirPilot Design*

While the original design aimed to implement a fully DRL-based controller without PID components, subsequent experiments revealed significant challenges. The DRL agent required extensive training time (exceeding 10 hours on standard computing hardware) and introduced significant flight instability, likely due to the black-box nature of DRL algorithms during training. To address these issues, we created the AirPilot to combine the simplicity and reliability of traditional PID control with the adaptability and optimization strengths of DRL (Fig. 7), PPO specifically due to its unique advantages discussed in the early session. This hybrid approach ultimately proved to be highly effective in achieving stable and responsive flight control. Given that our approach does not require any modifications to the PX4 autopilot flight controller, it can be seamlessly integrated with any drone utilizing the PX4 flight control software. This adaptability ensures broad applicability across various UAV platforms, enhancing the versatility and ease of deployment of our DRL-enhanced control system.

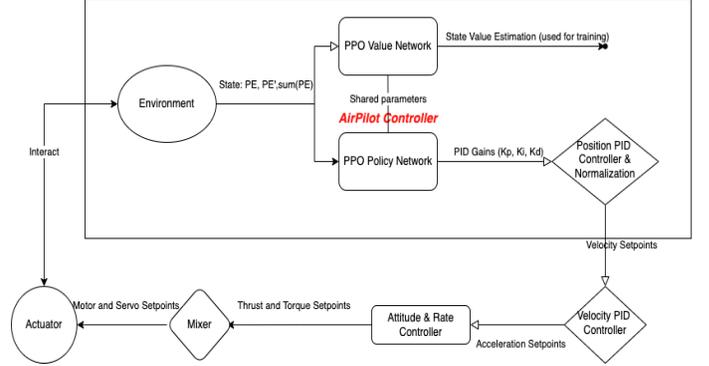

Figure 7. AirPilot controller architecture and its integration with the PX4 Autopilot controller. The output velocity setpoints from the AirPilot controller are sent to the velocity PID controller to generate the acceleration setpoints, which are then passed through the attitude and rate controllers to calculate the thrust and torque setpoints. Finally, the mixers send out the motor and servo setpoints for the drone to interact with the environment.

The PPO algorithm is imported from the stable_baselines3 library with a default 3e-4 learning rate, 64 batch size and 0.99 discount factor. The policy network and the value network contain two layers sharing the same parameters, each containing 64 neurons, but they use different heads for their respective tasks. As seen in Fig.7, the AirPilot policy network takes the drone's current PositionError, the derivative of PositionError, and the integral of PositionError as input and outputs the PID gains ($K_p, K_I, K_D$) as a nonlinear function of PositionError $PE$, its derivative $\frac{dPE}{dt}$ and integral $\int_t PE$. Using these gains, we modified the PID control laws as below (Eq.6) by adding nonlinearity to the PID controller and normalizing the velocity setpoints. $NormalizedV$ is set within [-1,1] (m/s) to prevent erratic behaviors (Eq.7).

$$V = K_p(PE, \frac{dPE}{dt}, \int_t PE)PE + K_D(PE, \frac{dPE}{dt}, \int_t PE)\frac{dPE}{dt}$$
$$+ K_I\left(PE, \frac{dPE}{dt}, \int_t PE\right)\int_t PE \quad (6)$$

$$NormalizedV = \frac{Velocity}{abs(Velocity) + 1} \quad (7)$$

The preliminary reward function we used for training AirPilot is designed as follows (Eq. (8), Eq. (9) and Eq. (10)):

$$Distance = norm(TargetPosition - StartingPosition) \quad (8)$$
$$EffectiveSpeed = Distance / (0.04 * (Timestep - 50)) \quad (9)$$
$$Reward = e^{EffectiveSpeed} * 10 \quad (10)$$

Where the $Distance$ is defined as the absolute distance between the target and the starting position. For simplicity and demonstration purpose, we define $EffectiveSpeed$ as the main optimization goal, combining the navigation speed, precision, overshoot, settling time and energy efficiency. It is calculated as $Distance$ divided by the time it takes for the drone to navigate from the starting point to the target, once the drone has stayed at the target stably for more than 50 timesteps (1 timestep = 0.04s). After the drone has stayed at the target stably for more than 50 timesteps, the next target will be randomly generated and a completely new PositionError will be calculated based on its current location and the next target. However, if the drone leaves the target within the 50 timesteps limit, the $Timestep$ is continuously increased to decrease the $EffectiveSpeed$, until it can stably stay at the target. This definition encourages large navigation speed, small settling time and small overshoot.

In our use case, the reward function is defined in a way that highly encourages precise navigation to the target, high effective speed and quick settling time by using the definitions above and an exponential function. The observation of any large angle of the drone body is also penalized heavily by a negative reward to prevent dangerous motions. Notably, after every 1000 timesteps, the episode is terminated to encourage an energy-efficient flight and find the shortest path from the starting point to the destination. After only 20,000 timesteps and 40 episodes of training (about 1.5 hours) on a normal computer, we obtain the fine-tuned policy from our AirPilot network. From the reward function, the DRL agent will evolve as we add more features, and it is much more advanced than a simple linear PID controller. The reward definition can also be easily modified towards the specific needs of other flight missions.

B. *3D A\* Algorithm*

Inspired by the 2-dimensional A* algorithm, we have also proposed a new algorithm specifically designed for collision-free drone navigation in the 3-dimensional space (Fig. 8). Initially, the algorithm creates a grid of valid points by considering the space between the obstacles and the dimensions of the drone. Each point in this grid represents a potential position for the drone to navigate. Then, starting from the initial position and moving towards the goal, the algorithm evaluates neighboring points, considering both the distance from the starting point and the estimated distance to the goal (value heuristic). The algorithm iteratively selects the most promising point to explore next, updating the scores accordingly. By prioritizing points with lower scores, the algorithm efficiently explores the space until it reaches the goal, reconstructing the shortest collision-free path based on the recorded parent nodes. Finally, the path is visualized in the 3D environment, enabling the drone to navigate from the initial position to the desired destination in the shortest path while avoiding obstacles.

```python
def a_star(start, goal, grid):
    """A* algorithm implementation for 3D space with 27-connected points"""
    
    open_set = [(0, start)]  # Priority queue, (f_score, node)
    came_from = {}   # Track parent nodes
    g_score = {start: 0}  # Cost to reach a node
    f_score = {start: heuristic(start, goal)}  # Estimated total cost
    
    while open_set:
        current = heapq.heappop(open_set)[1]
        
        if current == goal:
            return reconstruct_path(came_from, current)
        
        for neighbor in get_neighbors(current, space):
            tentative_g_score = g_score[current] + 1  # Assuming uniform cost of 1
            if neighbor not in g_score or tentative_g_score < g_score[neighbor]:
                came_from[neighbor] = current
                g_score[neighbor] = tentative_g_score
                f_score[neighbor] = g_score[neighbor] + heuristic(neighbor, goal)
                heapq.heappush(open_set, (f_score[neighbor], neighbor))

path = a_star((0, 0, 0), (1, 2.3, 1), valid_points)
print(path)
```

Figure 8. 3D A* Drone Path Planner pseudo code.

**EXPERIMENT/ RESULTS:**

*Integration of Vicon*

We carried out both manual and autonomous flights without any external positioning system in the early stages. Experiments showed that it is critical to integrate the Vicon tracking system with the drone's Inertial Measurement Unit (IMU) to enhance navigation precision before deploying fine-tuned AirPilot on the real drone. In the absence of the Vicon position system, the drone exhibited significant instability, as evidenced by a persistent yawing of approximately 90 degrees and noticeable lateral drift (Fig.9). These observations, captured in a supplementary video [8], highlight the inadequacy of the drone's onboard sensors and control algorithms to maintain stable flight without external position feedback.

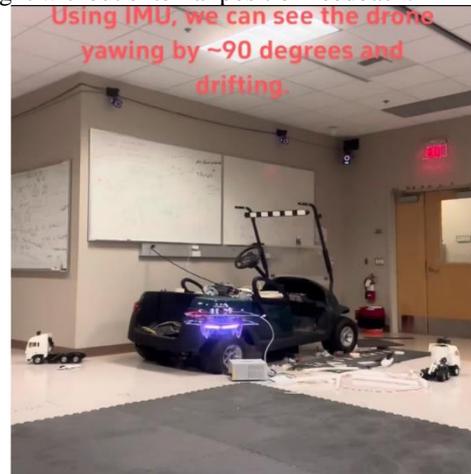

Figure 9. Unstable autonomous flight using IMU, without Vicon system.

After integrating the Vicon system, the flight stability of the drone increased significantly. The high-precision position and orientation feedback provided by the Vicon system enabled the drone to maintain a steady and controlled flight path, effectively eliminating the yawing and drifting observed in the absence of external position data. As seen in video [8], this enhanced stability allowed the drone to execute complex flight maneuvers with precision, completing tasks such as navigating through a series of waypoints (Fig.10). These experiments underscore the critical role of the Vicon system in ensuring precise control and maintaining the intended flight path.

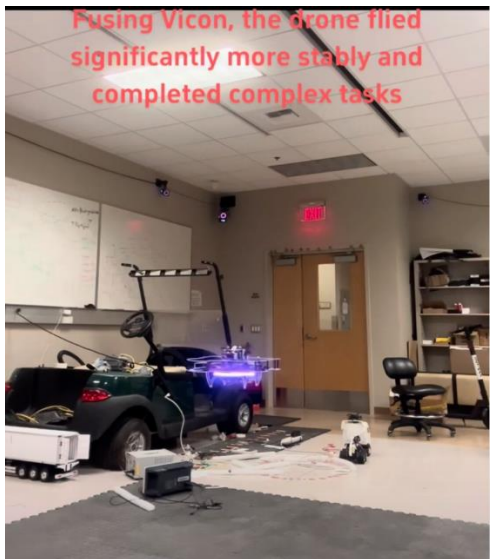

Figure 10. Autonomous flight with Vicon system.

### B. AirPilot in the Gazebo Simulator

Furthermore, with just a minimal amount of training (20,000 timesteps, spanning 40 episodes) on a standard computer, the implementation of advanced control algorithms—integrating Proximal Policy Optimization (PPO) with Proportional Integral Derivative (PID)—resulted in a significant enhancement in drone positioning accuracy within the simulator, achieving an impressive improvement of 90%. As illustrated in video [9], the use of a traditional PID controller to navigate the drone to the target coordinates (5, 5, 1.5) led to a substantial navigation error of 1.36 meters (Fig. 12). This was accompanied by an exceedingly long, nearly infinite, settling time, making the $EffectiveSpeed$ essentially 0 m/s, as the drone was unable to maintain a stable position (Fig. 11). In contrast, video [9] shows that our AirPilot controller drastically reduced the navigation error by 90%, bringing it down to just 0.14 meters (Fig. 13).

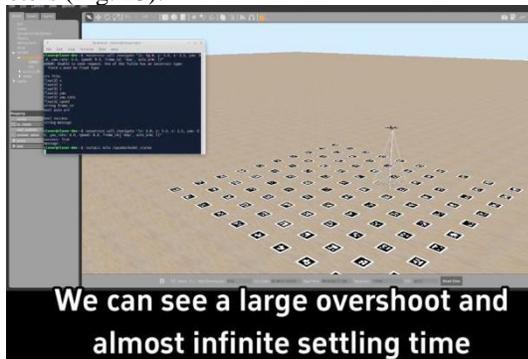

Figure 11. Navigating the drone to (5,5,1.5) using the PID controller in the simulator: large overshoot and infinite settling time.

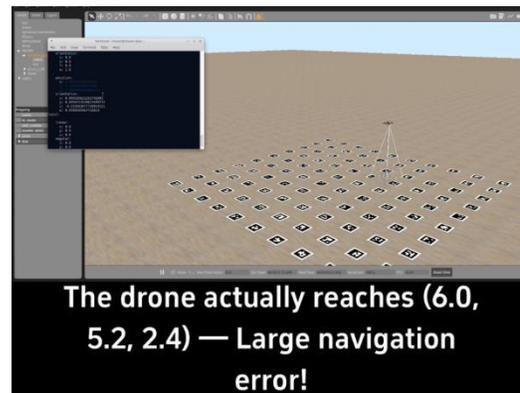

Figure 12. Navigating the drone to (5,5,1.5) using the PID controller in the simulator: large navigation error, large overshoot and infinite settling time.

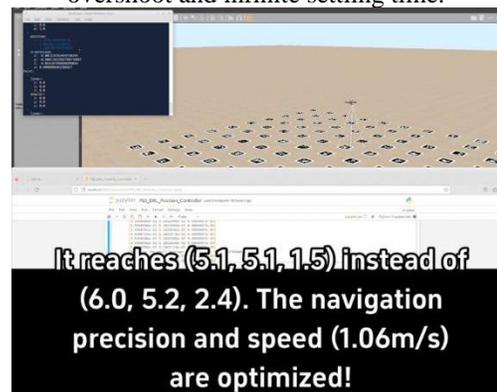

Figure 13. Navigating the drone to (5,5,1.5) using the DRL controller in the simulator.

As illustrated in Fig. 14, Fig.15, and Fig.16, there are a clear upward trend in $EffectiveSpeed$, downward trends in Settling Time and Overshoot, indicating the drone's enhanced navigation capabilities. This progression reflects the controller's growing efficiency and effectiveness in optimizing the drone's performance, including improvements in navigation speed, accuracy, and energy efficiency, as well as reductions in overshoot and settling time throughout the training process. These results align closely with the expectations established by the reward function. Worth mentioning, we also evaluated a pure PID controller using the steady-state PID gains identified by our AirPilot system, where the three PID gains were kept constant. The resulting $EffectiveSpeed$ was measured at 0.92 m/s with a settling time of 7.86s and an overshoot of 0.19 m, demonstrating that the AirPilot real-time adaptive controller outperforms its linear counterpart—a fine-tuned PID controller—by 21% in speed, 17% in settling time and 16% in overshoot respectively. This result further underscores the effectiveness of our adaptive approach in enhancing drone performance beyond what can be achieved with traditional PID control alone.

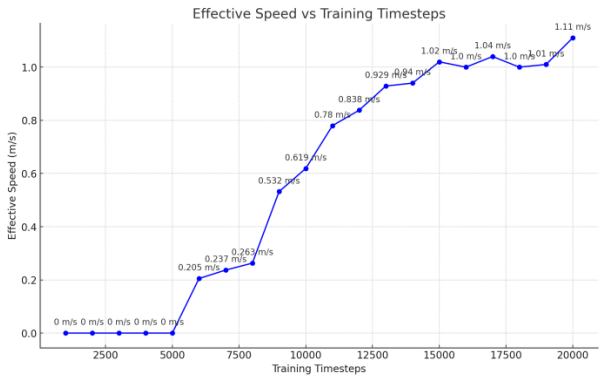

Figure 14. Using the AirPilot to navigate the drone, plot of Training Timesteps vs Effective Speed. Before 6000 Timesteps, the effective speed is 0m/s, because the drone fails to reach the goal.

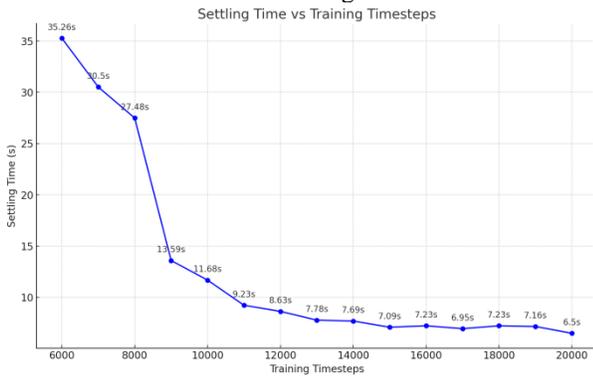

Figure 15. Using the AirPilot to navigate the drone, plot of Training Timesteps vs Settling Time. Before 6000 Timesteps, the settling time is infinite, because the drone fails to reach the goal.

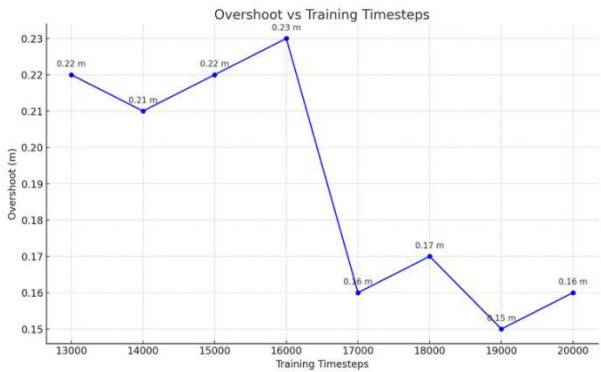

Figure 16. Using the AirPilot to navigate the drone, plot of Training Timesteps vs Overshoot. Before 13000 Timesteps, its overshoot is undefined, because the drone fails to pass the target before reaching it.

Fig.17 shows how PID gains learn to adapt to different Position Errors. As the done moves toward the target, the position error gradually decreases. For large position errors, $K_p$ is increased to provide a strong corrective action and drives the system quickly towards the setpoint. This helps in reducing the error rapidly. As the position error decreases and the system nears the setpoint, reducing $K_p$ can prevent overshoot and minimize oscillations, leading to a smoother convergence. When the error changes rapidly (e.g., when the drone moves towards the target at a high speed), $K_d$ is increased to help dampen the system's response, preventing overshoot and oscillations by anticipating the system's future behavior. $K_i$ is set to be a small constant to prevent overshooting due to accumulated corrective action, also indicating that the simulated drone navigation system is relatively accurate, and its steady-state error is small. This finding highlights the dynamic adaptability of the nonlinear AirPilot controller in real-time, showcasing its ability to optimize control parameters based on varying conditions, which is crucial for achieving precise and stable drone navigation. The ability to prevent overshoot and minimize oscillations through such adaptive behavior is critical for achieving stable and accurate flight control, demonstrating the potential for improved control in complex and dynamic environments.

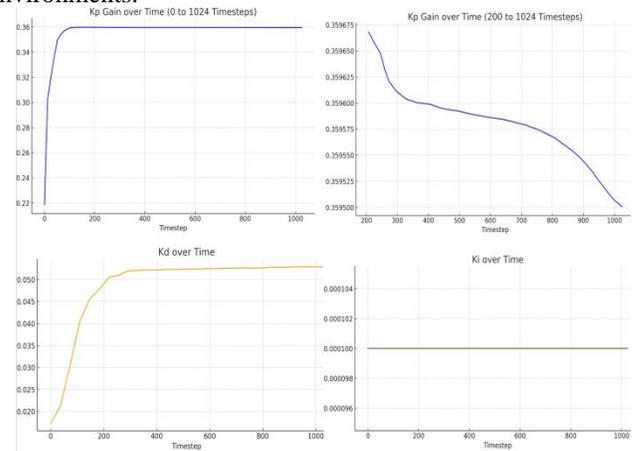

Figure 17. Using the AirPilot to navigate the drone to (5,5,1.5), Plot of PID Gains vs Time. As the drone approaches the target, $K_p$ first increases and then decreases, while $K_d$ keeps increasing and $K_i$ is a small constant.

The development of our AirPilot DRL-PID flight controller represents an early proof-of-concept, demonstrating the potential of Deep Reinforcement Learning (DRL) to meet the unique demands of various flight missions, such as enhancing effective speed, navigation accuracy, and settling time. While this work lays the groundwork for future research and innovation, it currently lacks well-defined metrics for systematically evaluating its performance due to the project timing constraints. Nevertheless, the principles demonstrated here can be extended beyond aerial vehicles to other autonomous systems, including robotic ground vehicles and marine robots, highlighting the broad applicability of DRL in autonomous control systems.

*C. AirPilot at the RCPSL*

We successfully implemented the AirPilot controller on a real COEX Clover drone by developing an innovative real-time interface between a personal computer, the Vicon motion capture system, and the drone's Raspberry Pi. This interface enables the seamless transmission of state estimations from the Vicon tracker to our DRL agent running on the personal computer. The DRL agent then computes the velocity setpoints, which are subsequently sent to the drone's Raspberry Pi for execution, ensuring precise and responsive control in real-time.

However, low-fidelity simulated sensors like image renderers often fail to reproduce the richness and noise produced by their real-world counterparts [9]. Due to the training being conducted in a simulated environment, and the well-documented Sim-To-Real challenge, also known as the reality gap [9], the drone exhibited less-than-ideal behavior during real-world deployment. As shown in the accompanying video [10], minor jittering and overshoot were clearly observed as the drone approached the target position in the real flight tests (Fig. 18). These qualitative observations underscore the challenges of deploying DRL-trained models in real-world environments and highlight areas for future refinement.

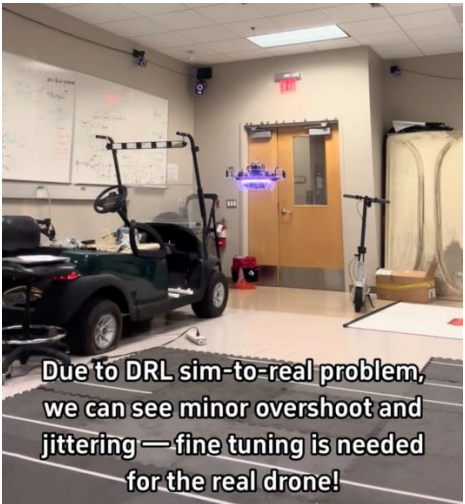

Figure 18. Navigating the Drone to (0,1,1) using the DRL controller in the real flight.

Provided that adequate training and fine-tuning are conducted in real-world flight conditions, the sim-to-real transfer challenge inherent in DRL does not compromise the overall superiority and learning potential of our sample-efficient DRL-based AirPilot drone controller. This robust performance establishes a strong foundation for further advancements in drone technology and autonomous navigation systems.

We believe our work to be among the pioneering efforts to deploy a DRL-enhanced drone flight controller in a real lab environment [11], rather than solely in simulation [12][13]. This achievement not only demonstrates the feasibility of applying DRL in real-world UAV operations but also serves as a steppingstone for future research in the field. Our results contribute to bridging the gap between simulation and real-world application, offering valuable insights for other researchers pursuing similar advancements.

*D. Testing 3D A\* Path Planner*

To ensure the drone navigates along a collision-free trajectory toward the target, we developed a three-dimensional A\* path planner that utilizes a heuristic approach. Our algorithm efficiently generates the shortest path in 3D space from any given starting point to the designated destination while avoiding all pre-known obstacles (Fig. 19). As shown in video [14], the generated navigation setpoints are transmitted to the drone at a frequency of 1 Hz, enabling it to safely and swiftly reach the target under the guidance of our AI-driven path planner (Fig. 20).

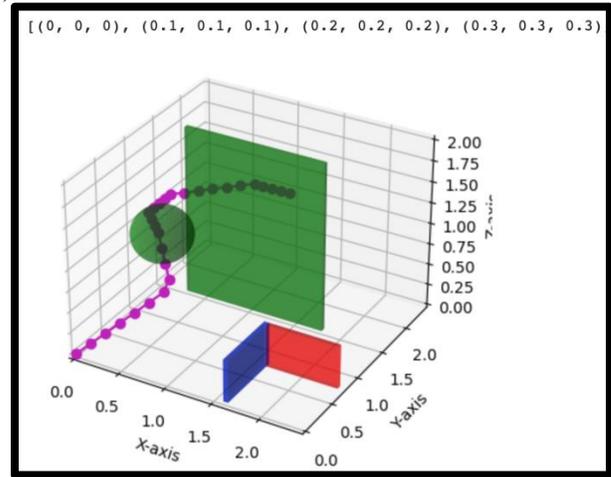

Figure 19. 3D A\* Path Planner can generate navigation setpoints from starting point to the target.

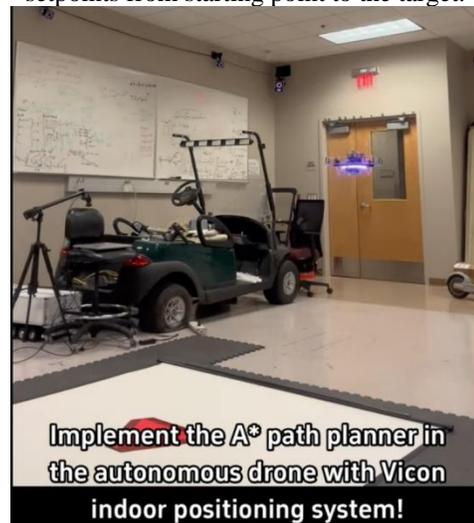

Figure 20. Deploying 3D A\* Path Planner in the Real Lab.

**CONCLUSION:**

In conclusion, this research introduces the AirPilot, a Deep Reinforcement Learning (DRL)-enhanced PID controller designed to improve UAV navigation in dynamic environments and to meet various customized mission-specific needs. By integrating Proximal Policy Optimization (PPO) with traditional PID control, AirPilot successfully addresses the limitations of linear PID controllers, offering enhanced precision, stability, adaptability and energy efficiency.

Testing in both simulated and real-world environments demonstrated significant improvements in navigation accuracy and stability, particularly when coupled with a high-precision Vicon tracking system. The integration of a 3D A\* path planner further ensured collision-free flight paths, enhancing mission success.

While challenges remain in transferring simulation-trained models to real-world applications, the AirPilot's success lays a strong foundation for future research and broader applications in autonomous systems, advancing the field of adaptive control in robotics.

**VII. ACKNOWLEDGEMENT**: